\documentclass{article}
\usepackage{arxiv}
\usepackage[utf8]{inputenc} 
\usepackage[T1]{fontenc}    
\usepackage{hyperref}       
\usepackage{url}            
\usepackage{booktabs}       
\usepackage{amsfonts}       
\usepackage{amsmath}
\usepackage{nicefrac}       
\usepackage{microtype}      
\usepackage{color}
\usepackage{orcidlink}
\usepackage{graphicx}
\graphicspath{ {./images/} }

\title{Subtraction-Based Tumor Segmentation and Lesion-Centered pCR Prediction for the MAMA-MIA Challenge}

\author{
 Kai Geissler\orcidlink{0009-0008-9709-0335} \\
  Fraunhofer Institute for Digital Medicine MEVIS\\
  Bremen, Germany\\
   \And
 Raphael Schäfer\orcidlink{0009-0000-4776-9333} \\
  Fraunhofer Institute for Digital Medicine MEVIS\\
  Bremen, Germany\\
}

\begin{document}
\maketitle
\begin{abstract}
We describe the submission of team FME to the MAMA-MIA Challenge, which evaluated primary tumor segmentation and prediction of pathological complete response (pCR) from pretreatment dynamic contrast-enhanced breast MRI on an external multi-country cohort. For segmentation, we trained a five-fold residual-encoder nnU-Net ensemble using only the first post-contrast minus pre-contrast image, combined with mirroring test-time augmentation and largest-connected-component filtering. For pCR prediction, we ensembled 25 pretrained 3D video classifiers trained on lesion-centred crops from the pre-contrast and first two post-contrast volumes. FME ranked second in both tasks. The segmentation method achieved a combined performance–fairness score of 0.882, with Dice 0.713 and normalized Hausdorff distance 0.099. The pCR method achieved a combined score of 0.664, balanced accuracy of 0.541, and equalized-odds disparity of 0.212. The results indicate that subtraction-based input and ensembling support robust tumor segmentation under cross-site domain shift, whereas pCR prediction from baseline DCE-MRI alone remains limited.

Submission repository: \href{https://github.com/FraunhoferMEVIS/MAMA-MIA-Challenge-FME}{github.com/FraunhoferMEVIS/MAMA-MIA-Challenge-FME} 
\end{abstract}

\keywords{Breast MRI \and Tumor Segmentation \and Pathologic Complete Response Prediction \and Deep Learning \and Algorithmic Fairness}

\section{Introduction}
The MAMA-MIA challenge (official title: Advancing Generalizability and Fairness in Breast MRI Tumor Segmentation and Treatment Response Prediction)~\cite{garrucho2026mama} at MICCAI 2025 (28$^{\text{th}}$ International Conference on Medical Image Computing and Computer Assisted Intervention) focused on advancing generalizability and fairness for two breast MRI image analysis tasks: primary tumor segmentation and prediction of pathological complete response (pCR) to neoadjuvant chemotherapy.
Models were developed on a public training cohort and evaluated on a private, multi-country test set from previously unseen sites.
Performance was summarized with combined scores that equally weighted task accuracy and fairness, the latter quantified as performance disparities across age, menopausal status, and breast density to encourage solutions that perform well on average and minimize systematic gaps across subpopulations.

Our submissions placed second in both tasks.
For segmentation, our pipeline used normalized subtraction volumes and a self-configuring nnU-Net ensemble with connected component post-processing and yielded accurate, stable masks under domain shifts with competitive fairness.
For pCR prediction, lesion-centered boxes with pretrained 3D classifiers achieved balanced accuracy above the baseline and a favorable fairness score on the hidden test set, despite the intrinsic difficulty of the task and low prediction performance for all teams.

This study details the public and private datasets, the challenge metrics, the modeling choices of team FME for both tasks, and a discussion of our official results.

\section{The MAMA-MIA challenge}
\subsection{Breast MRI data}
The MAMA-MIA challenge involved two breast MRI datasets: One public dataset for model training and another private dataset, that served as independent and hidden test data for the challenge. 

\paragraph{Public Training Data}
The public training data for model development consisted of 1,506 MRI volumes from 25 clinical centers in the United States~\cite{garrucho2025large}.
Its MRI volumes stemmed from four public datasets: Breast-MRI-NACT-Pilot~\cite{newitt2016single}, ISPY1~\cite{newitt2016multi}, ISPY2~\cite{li2022spy} and Duke-Breast-Cancer-MRI~\cite{saha2021dynamic}.
They were annotated by 16 radiologists who segmented all the primary tumors to form a joint dataset with coherent expert annotations. 
Of the 1,506 training cases, 440 (29.2\%) achieved pathological complete response, while 1051 (69.8\%) did not.
The label was missing for 15 (1.0\%) cases.
In addition, the training data contains metadata for each case, with information such as patient age, menopausal status, breast density and MRI scanner information.

\paragraph{Private Test Data}
The private test data consisted of 574 cases from three clinical institutions in Poland (30 cases), Lithuania (232 cases) and Spain (312 cases). 
Of the test cases, 160 (27.9\%) achieved pathological complete response, while 414 (72.1\%) did not.
During the challenge there was a validation leaderboard using 58 of these cases for metric computation while the final leaderboard was computed on all 574 cases.

\subsection{Challenge metrics}
The final ranking score for both tasks equally weights predictive performance ($S_p$) and subgroup fairness ($S_f$):
\begin{equation}
    S = \frac{1}{2}(S_p + S_f).
    \label{eq:combined_score}
\end{equation}
Fairness was assessed across age, menopausal status, and breast density.
For a fairness variable $v$ and metric $M$, let $\Delta_v(M)$ denote the difference between the maximum and minimum subgroup-average values of $M$.

\paragraph{Segmentation}
Segmentation performance combined the mean Dice similarity coefficient (DSC) and normalized Hausdorff distance (NormHD):
\begin{equation}
    S_p^{\mathrm{seg}} =
    \frac{1}{2}
    \left[
        \mathrm{DSC} + \left(1-\mathrm{NormHD}\right)
    \right].
    \label{eq:seg_performance}
\end{equation}
The segmentation fairness score was computed from subgroup disparities in DSC and NormHD:
\begin{equation}
    S_f^{\mathrm{seg}} =
    1 -
    \frac{1}{|\mathcal{V}|}
    \sum_{v \in \mathcal{V}}
    \frac{\Delta_v(\mathrm{DSC}) + \Delta_v(\mathrm{NormHD})}{2},
    \label{eq:seg_fairness}
\end{equation}
where $\mathcal{V}$ denotes the set of fairness variables. Higher values of both $S_p^{\mathrm{seg}}$ and $S_f^{\mathrm{seg}}$ are better.

\paragraph{pCR Prediction}
Classification performance was measured using balanced accuracy (BA) which can be expressed in terms of the true positive rate (TPR) and true negative rate (TNR):
\begin{equation}
    S_p^{\mathrm{cls}} =
    \mathrm{BA} =
    \frac{\mathrm{TPR} + \mathrm{TNR}}{2}.
    \label{eq:cls_performance}
\end{equation}
The equalized-odds disparity for a fairness variable $v$ was defined as
\begin{equation}
    D_{\mathrm{EO},v} =
    \Delta_v(\mathrm{TPR}) + \Delta_v(\mathrm{FPR}).
    \label{eq:eo_disparity}
\end{equation}
The classification fairness score was therefore
\begin{equation}
    S_f^{\mathrm{cls}} =
    1 - \overline{D}_{\mathrm{EO}} = 
    1 -
    \frac{1}{|\mathcal{V}|}
    \sum_{v \in \mathcal{V}} D_{\mathrm{EO},v}.
    \label{eq:cls_fairness}
\end{equation}
The official leaderboard reports the average equalized-odds disparity, $\overline{D}_{\mathrm{EO}}$, rather than the fairness score; hence, lower reported EOD values are better. The final classification score can be written as
\begin{equation}
    S^{\mathrm{cls}} =
    \frac{1}{2}
    \left[
        \mathrm{BA} +
        \left(1-\overline{D}_{\mathrm{EO}}\right)
    \right].
    \label{eq:cls_combined_score}
\end{equation}

\section{Methods}

\begin{figure}
    \centering
    \includegraphics[width=1.0\linewidth]{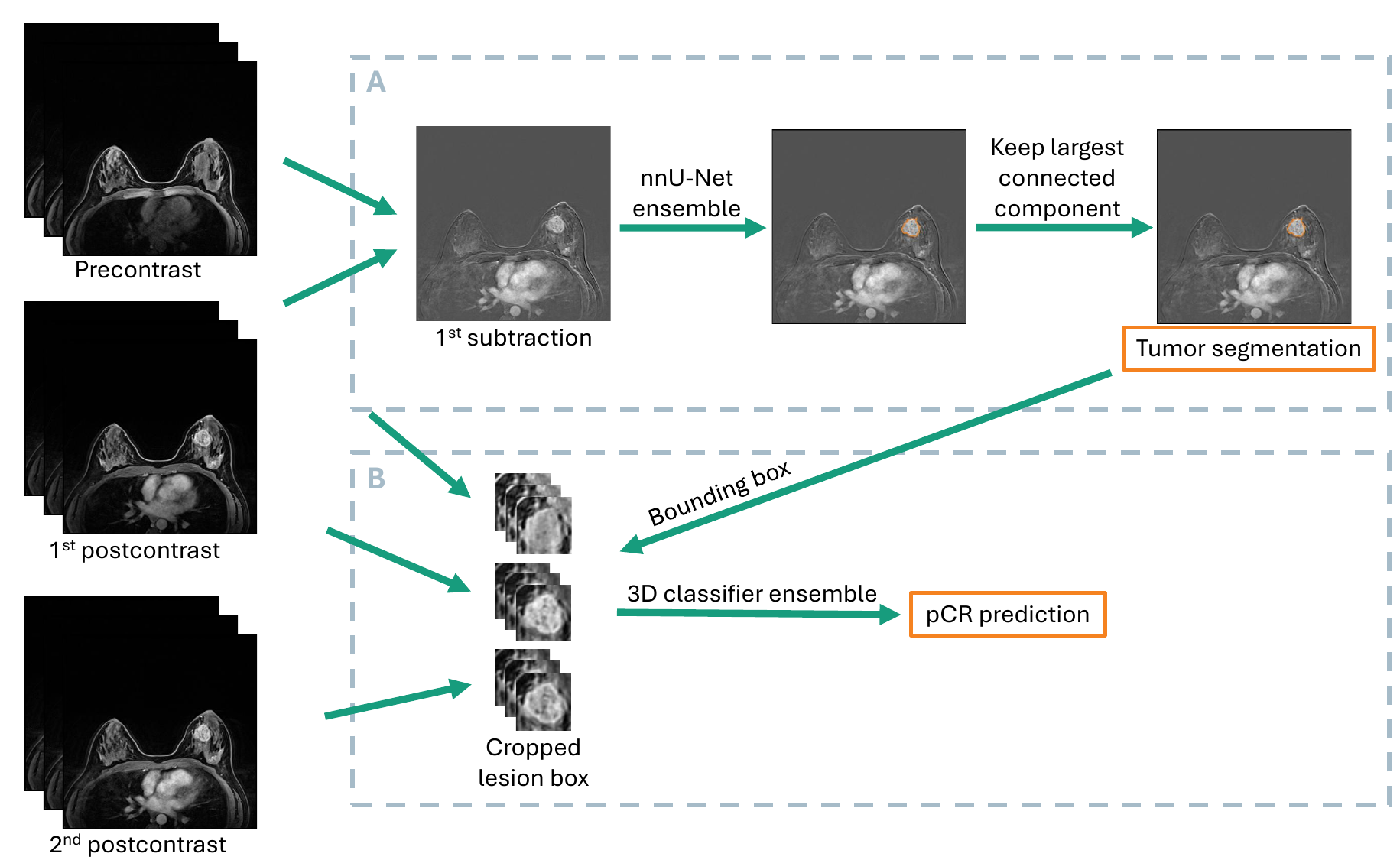}
    \caption{For segmentation (A), our method used a 5-fold ensemble of nnU-Nets on the 1$^\text{st}$ subtraction volume for segmentation. For pCR prediction (B), we used a 25-fold ensemble (comprising five selected hyperparameter configurations trained across five cross-validation folds) of TorchVision video classifiers on lesion-cropped bounding boxes of the first three DCE timepoints.}
    \label{fig:graphical_abstract_mama_mia}
\end{figure}

\subsection{Segmentation}
One of the big issues in dynamic contrast-enhanced MRI is that the timing of image acquisition varies between different clinical centers, and sometimes even within a single center. 
In the MAMA-MIA dataset, the lesion segmentations were created on the first post-contrast images.
To minimize the effect of different acquisition timings, we used only the first subtraction image (i.e. first post-contrast image minus pre-contrast image) as model input. 
\figurename~\ref{fig:graphical_abstract_mama_mia} (A) shows our segmentation pipeline.

As segmentation model we trained a 5-fold ensemble of nnU-Nets~\cite{isensee2021nnu} using the nnUNetPlannerResEncL plans~\cite{isensee2024nnu} with 24 GB of GPU memory and the 3d\_fullres model configuration. 
The nnU-Net framework takes care of preprocessing steps like resampling and intensity normalization using field-proven heuristics, so that we did not investigate further preprocessing steps.

During inference, mirroring test-time augmentation was activated.
As post-processing we conducted a connected component analysis to keep only the largest connected component in each case.

The resulting nnU-Net configuration used a voxel size of $0.7\times0.7\times2.0$~mm$^3$, a patch size of $256\times256\times80$ voxels and a batch size of 2.
Training was performed on an NVIDIA A100 GPU using PyTorch 2.7 and Python 3.11 in a CUDA 12.8 environment.

\subsection{Classification}
Our classification pipeline is shown in \figurename~\ref{fig:graphical_abstract_mama_mia} (B).
To predict pathological complete response (pCR), the first three DCE-MRI volumes of each case were selected as model input. 
We chose three volumes, to be able to leverage pretrained video classification models, that come with three input channels.
The slice-dimension of the MRI inputs was fed to the temporal dimension of the video classification models.
All three volumes were z-score normalized using the same mean and standard deviation, which were computed only on the pre-contrast volume.
To reduce the region-of-interest in classification, the bounding box around each primary lesion was cropped from the input volume using the reference segmentation. 
During inference we used the predicted segmentation masks to crop the lesions.
The cropped boxes were resampled to a fixed image size, which differed based for each specific model trained.

We employed 3D classification models from the collection of TorchVision video models~\cite{torchvision2016}, which are deep learning models pretrained for video classification.
Hyperparameter search with Ray Tune~\cite{liaw2018tune} was used to select the model hyperparameters using 5-fold cross-validation and a simple random sampling search strategy.
\tablename~\ref{tab:mama_mia_hyperparameter_search_space} shows the search space.
We selected five sets of five folds which performed best on average with respect to the re-implemented ranking score and balanced accuracy, resulting in a final ensemble of 25 models.

\begin{table}
  \centering
  \caption{Hyperparameter search space for lesion classification model training.}
  \begin{tabular}{lr}
  \toprule
  Hyperparameter & Search Distribution \\
  \midrule
  Learning rate & Loguniform(1e-6, 1e-3) \\
  Final learning rate & Loguniform(1e-8, 1e-4) \\
  Momentum / $\beta_1$ & Uniform(0.8, 0.98) \\
  Weight decay & Loguniform(1e-6, 1e-1) \\
  Batch size & Uniform(16, 32) \\
  Label smoothing & Loguniform(1e-4, 1e-1) \\
  x-y-Resolution & Uniform(50, 150) \\
  z-Resolution & Uniform(16, 50) \\
  Model architecture & {r3d\_18, swin3d\_t, r2plus1d\_18, mc3\_18} \\
  \bottomrule
  \end{tabular}
  \label{tab:mama_mia_hyperparameter_search_space}
\end{table} 

The model used a binary cross entropy loss and the AdamW optimizer~\cite{loshchilov2017decoupled} with a cosine annealing learning rate scheduler~\cite{loshchilov2016sgdr}.

For data augmentation we used the \texttt{batchgeneratorsv2} library, which offers the same functionality as the \texttt{batchgenerators} library~\cite{isensee2020batchgenerators}, but has a stronger focus on augmentation runtime. 
To speed up model training we used automatic mixed precision, which automatically reduces the floating point precision of certain deep learning operations to speed up model training and reduce the memory footprint~\cite{micikevicius2017mixed}. 
The models were trained for a maximum of 25 epochs and we used early stopping based on the ranking score computed on the validation folds. 
The ranking score re-implemented the task 2 ranking score from the MAMA-MIA challenge using the age and menopausal status metadata. 
The breast density metadata was not used because it was missing in almost all training cases. 

Training was performed on an NVIDIA RTX A5000 GPU using PyTorch 2.7 and Python 3.11 in a CUDA 11.8 environment.

During inference, test-time augmentation was applied by flipping the input volume once along each axis, resulting in 8 inference passes for each volume.
All the prediction results, both from the ensemble and test-time augmentation, were averaged to arrive at the final prediction which used a classification threshold of 0.5.

\section{Results}
\subsection{Segmentation}
\tablename~\ref{tab:mama_mia_segmentation_leaderboard} shows our placement within the final challenge leaderboard for task 1 (segmentation).
In total there were 21 submissions, 12 of which outperformed the baseline model, which was a 3D nnU-Net. 
Our submission (team FME) achieved the second place with a combined score of 0.8820.
The first three places were close to each other, achieving combined scores of 0.8858, 0.8820 and 0.8782. 
The same holds for the fairness scores (0.9531, 0.9574, 0.9482) and the performance scores (0.8185, 0.8066, 0.8083). 
While the first place submission had a better Dice score (0.7360 vs. 0.7125 and 0.7182) than the other places, the normalized Hausdorff distance (0.0990, 0.0993, 0.1017) and fairness scores (0.9531, 0.9574, 0.9482) were much closer to each other for the three top performing submissions.

\begin{table}
\centering
\caption{Selected entries of official final MAMA-MIA leaderboard for task 1 (segmentation) as published in~\cite{garrucho2026mama}. Our submission is team FME.}
\label{tab:mama_mia_segmentation_leaderboard}
\footnotesize
\begin{tabular*}{\textwidth}{@{\extracolsep\fill}clccccc}
\toprule
 &  & Combined & Fairness & Performance & &  \\
Rank & Team & {Score $\uparrow$} & {Score $\uparrow$} & {Score $\uparrow$} & {DSC $\uparrow$} & {NormHD $\downarrow$} \\
\midrule
 1 & MIC & 0.8858 & 0.9531 & 0.8185 & 0.7360 & 0.0990 \\
 2 & \textbf{FME} & 0.8820 & 0.9574 & 0.8066 & 0.7125 & 0.0993 \\
 3 & ViCOROB & 0.8782 & 0.9482 & 0.8083 & 0.7182 & 0.1017 \\
 \ldots & \ldots & \ldots & \ldots & \ldots & \ldots & \ldots \\
13 & \textit{Baseline} & 0.8290 & 0.9373 & 0.7208 & 0.6871 & 0.2455 \\
\ldots & \ldots & \ldots & \ldots & \ldots & \ldots & \ldots \\
\bottomrule
\end{tabular*}
\end{table}

\subsection{Classification}
\tablename~\ref{tab:mama_mia_classification_leaderboard} shows our placement on the final leaderboard for task 2 (classification). 
In total, 15 teams made submissions, only three of which outperformed the random classifier baseline with respect to the combined score, with our team (FME) reaching the second place.
The top three combined scores were 0.6907, 0.6642 and 0.6625.
While the top 1 submission had an equalized-odds disparity of 0.1150 and a balanced accuracy of 0.4964, our second place had an equalized-odds disparity of 0.2121 and a balanced accuracy of 0.5405.
The third place had an equalized-odds disparity of 0.2013 and balanced accuracy of 0.5263.

\begin{table}
\centering
\caption{Selected entries of official final MAMA-MIA leaderboard for task 2 (classification) as published in~\cite{garrucho2026mama}. EOD is the equalized odds disparity (i.e. 1 $-$ fairness score). Our submission is team FME.}
\label{tab:mama_mia_classification_leaderboard}
\begin{tabular*}{\textwidth}{@{\extracolsep\fill}clcccc}
\toprule
 & & {Combined} & {} & {Balanced} & {} \\
{Rank} & {Team} & {Score $\uparrow$} & {EOD $\downarrow$} & {Accuracy $\uparrow$} & {AUC $\uparrow$} \\
\midrule
 1 & pimed-lab   & 0.6907 & 0.1150 & 0.4964 & 0.4556 \\
 2 & \textbf{FME} & 0.6642 & 0.2121 & 0.5405 & 0.5722 \\
 3 & AI Strollers & 0.6625 & 0.2013 & 0.5263 & 0.5218 \\
 4 & \textit{Baseline} & 0.6431 & 0.2179 & 0.5042 & 0.5031 \\
 \ldots & \ldots & \ldots & \ldots & \ldots & \ldots \\
\bottomrule
\end{tabular*}
\end{table}

\section{Discussion}
\subsection{Segmentation}
Our segmentation strategy was to train a 5-fold nnU-Net ensemble on the first subtraction image and apply simple connected-component post-processing which proved to be both effective and robust across the hidden, multi-country test sites. 
Despite notable domain shifts between the public US training cohort and the private test cohort from Poland, Lithuania, and Spain, the model achieved a combined score comparable to the top entries (\tablename~\ref{tab:mama_mia_segmentation_leaderboard}), with only minor absolute differences among the first three teams. 
We hypothesize that the use of only the first subtraction image to mitigate acquisition timing variability, ensembling, and conservative post-processing supported robust cross-site performance.

Performance and fairness moved largely in tandem for the top methods.
While the first-place model reported a slightly higher Dice, the normalized Hausdorff distances and fairness scores among the top entries were tightly clustered, indicating that no participating team had a fundamentally superior method compared to the others.
In our approach, largest-component filtering reduced small spurious detections, improving boundary metrics and contributing to better fairness by lowering group-specific error variability.
A trade-off is that such filtering may undersegment multifocal disease in some cases; however, it turned out not to hurt our segmentation performance in this challenge setting.

From an engineering standpoint, limiting inputs to the first post-contrast minus pre-contrast image simplified modeling and avoided dealing with site-specific multi-phase timing differences documented in DCE breast MRI.
As a consequence, the ensemble focused on the lesion morphology, spatial localization and boundary delineation rather than kinetic curve modeling.
Future gains may come from harmonization and multi-phase strategies that safely reintroduce kinetics without re-amplifying timing variability.

We did not explicitly optimize for fairness beyond the robustness measures mentioned above.
Nonetheless, our fairness score indicates that generalization-enhancing choices can reduce disparities even without group-aware training.
It should be noted though, that the fairness score of all methods in the top 12 were close (0.934--0.962) and the differentiator was the performance score (0.736--0.819).
Dedicated fairness methods, such as group-balanced sampling, per-group calibration, or adversarial debiasing, might help improve equalized performance across age, menopausal status, and breast density.

\subsection{Classification}
In contrast to segmentation, pCR prediction from baseline DCE-MRI proved challenging for all teams, with balanced accuracies close to chance and modest differences in combined scores (\tablename~\ref{tab:mama_mia_classification_leaderboard}).
Our approach to crop around the primary lesion and use pretrained 3D video classifiers over the first three volumes yielded a balanced accuracy above the baseline but still limited in absolute terms.
But although we achieved a balanced accuracy of 0.5405 and the highest reported AUC among participating methods (0.5722), our balanced accuracy was only marginally different from random prediction (p = 0.063) in the official analysis.
This aligns with the broader literature: pCR is driven by biological and microenvironmental factors that are only partially reflected in baseline morphology and early enhancement patterns \cite{li2024breast,khan2022deep}. 
Without richer kinetics, diffusion, T2-weighted information, or clinical covariates (e.g., receptor status, molecular subtype), the signal available to a purely image-based classifier is weak.

We observed similar classifier performance across architectures (2D, 2D with 3D attention, 3D) in preliminary experiments, likely due to the limited data scale and label structure given the difficulty of the task. 
Our hyperparameter search, test-time augmentation, and ensemble selection improved generalization but did not overcome fundamental signal constraints.
Importantly, we could not leverage breast density metadata during training because it was largely missing in the public dataset; this limited our ability to optimize fairness. 
The fairness metric, equalized odds disparity across age, menopausal status, and breast density, penalizes subgroup differences in both true positive rate and false positive rate at the operating point. 
In small, imbalanced cohorts with domain shift, threshold selection and calibration can contribute substantially to disparities.
Group-aware calibration (e.g., temperature scaling per subgroup), reweighting, or fairness-aware postprocessing (e.g., equalized odds adjustments) may reduce disparities without sacrificing overall performance, but they require reliable metadata and sufficient subgroup support.

\section{Conclusion}
In this study we described the approach of team FME for the segmentation and classification tasks of the MAMA-MIA challenge which won the second place in both tasks and was the only submission that made it in the top 3 of both tasks.
In summary, the challenge highlights a practical contrast: segmentation generalizes well with robust preprocessing and strong defaults, while baseline-examination-only pCR prediction from DCE-MRI remains fundamentally data- and modality-limited. 
Our results suggest that future progress might depend less on classifier architecture and more on comprehensive imaging and further clinical inputs, calibrated decision policies, and fairness-aware training pipelines supported by complete and consistent metadata.

\section*{Conflict of Interest}
The authors have no competing interests to declare that are relevant to the content of this article.

\bibliographystyle{unsrturl}  
\bibliography{references}

\end{document}